\documentclass[journal,twoside]{IEEEtran}
\usepackage{cite}

\title{Applications of Generative AI in Healthcare: algorithmic, ethical, legal and societal considerations}
\author{Onyekachukwu Rhema Okonji - onyekaokonji@gmail.com \\ Kamol Yunusov - lawrium@hotmail.com\\ Bonnie Gordon - bonniegordon2610@gmail.com}

\begin{document}

\maketitle

\begin{abstract}
    Generative AI is rapidly transforming medical imaging and text analysis, offering immense potential for enhanced diagnosis and personalized care. However, this transformative technology raises crucial ethical, societal, and legal questions. This paper delves into these complexities, examining issues of accuracy, 
    informed consent, data privacy, and algorithmic limitations in the context of generative AI's application to medical imaging and text. We explore the legal landscape surrounding liability and accountability, emphasizing the need for robust regulatory frameworks. Furthermore, we dissect the algorithmic challenges, 
    including data biases, model limitations, and workflow integration. By critically analyzing these challenges and proposing responsible solutions, we aim to foster a roadmap for ethical and responsible implementation of generative AI in healthcare, ensuring its transformative potential serves humanity with utmost care and precision. 
\end{abstract}


\section{Introduction}
The landscape of medical diagnosis is undergoing a seismic shift. Generative AI, with its ability to analyze images and craft insightful text reports, holds immense promise for revolutionizing healthcare. Yet, beneath the surface of this technological marvel lie ethical, societal, and legal complexities that demand careful scrutiny.

This report embarks on a critical exploration of this transformative technology. We delve into the ethical considerations that underpin its application in medical imaging and text analysis, examining the delicate balance between accuracy, informed consent, and data privacy. We navigate the uncharted legal landscape, 
dissecting the challenges of liability and accountability in the face of AI-driven diagnoses. Finally, we turn a discerning lens to the algorithmic intricacies, highlighting concerns around data biases, model limitations, and the need for seamless workflow integration.

This is not merely an academic exercise. It is a call for responsible innovation, a plea to harness the power of AI while safeguarding the well-being of patients. By dissecting the complexities that lie at the heart of generative AI in medicine, we aim to pave the way for its ethical and responsible implementation, 
ensuring that this transformative technology serves humanity with the utmost care and precision. 

\section{Selection of application}
This report will be looking at 4 critical aspects related to the application of Generative AI models in healthcare with a focus on medical imaging and text.

Generative AI's burgeoning influence in medical imaging and text analysis demands a rigorous exploration of its ethical, societal, and legal implications. This report delves into the intricacies of this potent technology, dissecting its promising potential alongside the complexities that necessitate careful consideration.
Ethical questions weave throughout the fabric of generative AI in medicine. Accuracy, a cornerstone of trust, necessitates rigorous testing and transparency about performance limitations. Informed consent becomes paramount when AI generates potentially life-altering reports, demanding clear communication about its 
role and limitations. Accountability rests not only with developers but also with healthcare institutions wielding this tool, requiring robust audit trails and clear channels for redressal. Data privacy, ever-sensitive in healthcare, demands stringent safeguards against unauthorized access and secondary use. 

Societal considerations matter too. Will generative AI exacerbate existing healthcare disparities? How can we ensure equitable access to this technology for all? What are the public's concerns about AI in healthcare? How can we build trust and address ethical concerns to ensure widespread adoption? How will generative 
AI impact healthcare professionals? What are the implications for job displacement and the need for retraining and reskilling? 

Legally, Generative AI navigates uncharted waters. While existing regulations apply to data privacy and medical practice, specific frameworks for AI-generated diagnoses and reports are nascent. Issues of negligence and liability in the event of AI-based errors will require careful legal dissection, balancing accountability 
with fostering innovation. Ensuring transparency and explainability of AI algorithms becomes crucial for informed legal decisions and building public trust.

Algorithmic considerations cast an introspective lens on generative AI's limitations. Concerns about insufficient training data loom large, as biases within this data can lead to discriminatory outputs. Hallucinations and inaccuracies in AI-generated images or text can have dire consequences, necessitating stringent filters 
and human oversight. Training models and running complex algorithms necessitate substantial computational resources, posing challenges for wider adoption in resource-constrained settings. Moreover, integrating AI seamlessly into existing clinical workflows requires careful consideration, ensuring minimal disruption and smooth 
human-machine collaboration. The real-time nature of medical decision-making further complicates matters, demanding swift AI inference and processing without compromising accuracy.

This report aims to untangle this intricately woven tapestry of ethical, societal, and legal threads, illuminating both the transformative potential and the challenges inherent in generative AI's application to medical imaging and text. By fostering a comprehensive understanding of these complexities, 
we can pave the way for responsible innovation, harnessing AI's power to serve humanity while safeguarding the well-being of patients and upholding the highest ethical standards. 

\section{Analysis}

\subsection{Ethical aspects}
Artificial Intelligence (AI) has revolutionized the field of medical imaging, bringing forth numerous advancements and opportunities. 
However, with these advancements come important ethical considerations that must be addressed. This report explores the ethical aspects of AI 
in medical imaging, focusing on privacy, bias, and accountability.  AI has the potential to improve medical imaging by making them more efficient, affordable, and widely available.\cite{e1}  
With the use of Generative AI in medical imaging there has been as increase in the precision on diagnosis, treatment care plans and monitoring of numerous diseases. 
The use of specific techniques such as generative adversarial networks (GANs) and variational autoencoders (VAEs) strengthen medical imaging by generating synthetic images and 
improving reconstruction.\cite{e2} However, this is only going to be feasible if patients and medical professionals agree that AI medical devices (AIMDs) are reliable and trustworthy\cite{e3}.  
Ethical considerations must be made when using LLMs within medical radiography. While these models yield promising results, there are elements that must be considered when using these models.
\subsubsection{Privacy}
AI in medical imaging relies heavily on patient data, which raises concerns about privacy and data security. It is crucial to ensure that patient information is protected and used 
responsibly. As these models rely extensively on large datasets such as medical records, genetics and ethnical information, stricter regulations and robust security measures should be 
implemented to safeguard patient privacy and prevent unauthorized access or misuse of data\cite{e4}. Severe consequences of unauthorised use or breaches within the data security can result in 
identity theft, privacy violations and misuse of personal information.
\subsubsection{Bias}
AI algorithms used in medical imaging have the potential to introduce biases that may impact healthcare outcomes. It is essential to address and mitigate these biases to ensure fairness and 
accuracy in diagnosis and treatment. Transparency in algorithm development, diverse training datasets, and continuous monitoring can help identify and rectify biases in AI systems.
\subsubsection{Transparency and Accountability}
As AI systems become more integrated into medical imaging, questions of accountability arise. Who is responsible when an AI system makes a wrong diagnosis or misses a critical finding? 
Clear guidelines and mechanisms should be established to assign accountability and address any errors or adverse outcomes resulting from AI-assisted diagnosis\cite{e6}. Additionally, healthcare 
professionals should be trained to understand the limitations and potential biases of AI systems. 
\subsubsection{Autonomy of use}
Some concerns have stemmed from the impact on patient-doctor rapport and patient autonomy. Whilst LLMs provide effective insights and support to healthcare professionals, 
the over-reliance on the AI system can cause a wedge between patients trusting their care providers and lack of human touch. Patient relationship is built on empathy, care and 
understanding so to maintain such level of care, healthcare professionals need to use LLMs simply as a tool in supporting their decisions or care plan with the involvement of the patient.\cite{e7}  

\subsection{Societal aspects}
The advent of artificial intelligence (AI) technologies has revolutionized many sectors, including healthcare. Particularly, as mentioned, AI can be instrumental in medical imaging, 
enabling precision and accuracy in detecting and diagnosing diseases. However, this increased reliance on AI may impact the nature of social interaction in healthcare.
\subsubsection{Human touch}
Healthcare has traditionally been a high-touch field, with human interaction playing a critical role in patient care. The introduction of AI, while augmenting diagnostic efficiency, 
could potentially alter this dynamic. It may reduce direct interaction between healthcare providers and patients, as AI takes over tasks such as image interpretation and diagnosis. 
This could lead to a shift in the patient-provider relationship, posing questions about the importance of human touch and interaction in the healing process.\cite{s4} According to a study by 
Huang and Liang (2020), the potential lack of human touch in AI-based services may affect patient satisfaction and the perceived quality of care.\cite{s12} They argue that while AI improves 
efficiency, it may not be able to replicate the empathy, reassurance, and personal care that human healthcare providers offer.  Patients often derive comfort from the human interaction 
they receive during their healthcare experience, which AI, at its current state, cannot provide. This lack of 'human touch' could lead to a perceived decrease in the quality of care 
and may affect patient satisfaction. The implementation of AI in healthcare settings also brings forth complex societal issues that need addressing. These issues span the spectrum 
from ethical considerations, such as informed consent and privacy concerns, all of which will be discussed in this paper, to social implications, like potential job displacement and the digital divide.\cite{53}
\subsubsection{Job displacement}
From a societal perspective, the increased use of AI could potentially displace jobs, particularly those that involve routine tasks. AI can automate certain tasks, freeing up radiologists and other 
imaging professionals to focus on areas requiring human expertise. For example, AI algorithms can be trained to accurately identify abnormalities in imaging scans, thus improving efficiency and 
reducing the workload of clinicians.\cite{s29} However, AI is also likely to create new roles within the field. The design, development, deployment and maintenance of AI systems will require new skills, leading 
to the creation of jobs such as AI specialists and data scientists.\cite{s25} This necessitates a shift in the education and training of healthcare workers. Reskilling will be key in helping these professionals 
adapt to the changing landscape. Medical education will need to incorporate AI literacy, including understanding the strengths and limitations of AI, data interpretation, and ethical considerations.\cite{s7} 
Therefore, continuous professional development programs will need to focus on skill acquisition in AI-related areas. Furthermore, interdisciplinary collaboration between medical professionals, computer 
scientists, and engineers will become increasingly important.\cite{s8} Efforts should also focus on developing AI applications that complement, rather than replace, the human aspects of healthcare. 
This could involve creating AI systems that still involve human healthcare providers in the decision-making process, or developing AI tools that assist healthcare providers in delivering more personalised care.\cite{s7}
\subsubsection{Digital divide}
Moreover, there's a risk of exacerbating the digital divide, as access to AI-powered healthcare may be limited to certain segments of society, potentially widening health disparities. It is accepted that 
AI algorithms can help analyse medical images more efficiently and accurately, potentially reducing diagnostic errors, enhancing patient care, and improving health outcomes.\cite{s7} 
However, if not properly managed, AI could also exacerbate health disparities. It's important to note that AI algorithms are often trained on datasets that might not be representative of diverse populations. 
If these datasets predominantly include data from certain groups and exclude others, this could lead to biased outcomes and widen existing health disparities.\cite{s18} Additionally, AI can potentially improve access 
to care in underserved areas by allowing for remote diagnostics and consultations, thus overcoming geographical barriers.\cite{s3} However, this opportunity also presents a risk: if AI advancements are primarily 
accessible to affluent communities due to cost or infrastructure requirements, it could increase health inequalities. Addressing health disparities in AI requires a careful consideration of legal frameworks. 
Issues such as privacy, data security, and informed consent are key.\cite{s28} Furthermore, regulatory mechanisms must ensure the quality and safety of AI systems and their algorithms. Anti-discrimination laws 
must also be enforced to prevent biased data collection and usage.\cite{42}
\subsubsection{Public trust}
Lastly, while social aspect of AI is analysed, consideration of public trust is a crucial factor for the adoption of AI in healthcare, including in the realm of medical imaging. AI can potentially revolutionize 
healthcare by making diagnoses more accurate and efficient, but this relies on public acceptance and trust.\cite{s21} With the increasing use of AI in medical imaging, concerns about privacy, reliability, and fairness are paramount.\cite{s15}

Public trust is the bedrock upon which the successful implementation and acceptance of AI applications in healthcare are built.\cite{s1} Transparency about how AI is used in medical imaging is vital to gain and maintain public trust. 
This involves clear communication about the AI decision-making process, its accuracy, and potential limitations.\cite{33} Regulations that protect patient interests are also important. This includes laws to safeguard patient data, 
ensure the reliability and validity of AI algorithms, and uphold the principles of fairness and equity.\cite{s11} Strong regulations thus serve as a guarantee to the public that their interests are protected, further fostering trust.\cite{33}
Societal frameworks play a fundamental role in fostering public trust in AI use in healthcare. These frameworks include educational initiatives to improve public understanding of AI, and mechanisms for public input and feedback 
on AI use.\cite{s1} Complex societal issues such as inequality and bias in AI algorithms must also be addressed. For instance, AI systems trained on data from one demographic might not work as well for another, leading to disparities 
in healthcare outcomes.\cite{s11} Policies should be in place to ensure diversity in training data and to test AI systems across different groups.\cite{33}

For a more comprehensive detail on the aforementioned points, please visit Appendix A.

\subsection{Legal aspects}
\subsubsection{Accuracy}
Although AI is considered to have the potential to improve medical efficiency and precision, especially in medical imaging, its reliance on the quality of training data can lead to diagnostic errors, which could potentially harm patients. 
In conventional practices, incorrect diagnoses can result in legal liabilities, with the responsibility usually resting on the healthcare providers. However, in the context of AI, an important question emerges: 
"Who is accountable for AI negligence - the healthcare providers, the developers, or the institutions using AI?"\cite{31} Considering AI's self-governing decision-making and learning capabilities could reshape conventional 
legal notions of liability and accountability, it is debated that distinctive legal frameworks need to be established.\cite{32} From a legal viewpoint, accuracy isn't solely about a correct diagnosis; it also relates to the rights and dignity 
of those availing these services.

\subsubsection{Informed consent}
One potential solution to eliminate this issue is to provide patients with an informed choice, making them aware of potential risks and benefits of proposed medical procedures.\cite{34} Based on this informed consent, patients have the autonomy 
to make their own decisions, a principle that is legally mandated in healthcare systems globally.\cite{35} The significance of informed consent escalates when patient data is processed by AI,\cite{31} since decisions or recommendations made by AI systems 
based on this data can significantly impact patient health, underscoring the critical nature of informed consent.\cite{37} However, the concept of informed consent poses a considerable challenge due to the complexities of understanding potential 
risks and benefits of AI integration in their medical procedures and the "decision making" process of AI. This lack of understanding could potentially undermine the legitimacy of informed consent.\cite{38} Understanding and defining the "decision making" 
process also brings forth complex legal concerns. This requires addressing issues such as data protection, transparency, and accountability.\cite{79} However, the "black box" nature of AI, where decision-making processes are not easily understood, 
adds another layer of complexity.\cite{40} Therefore, incorporating data protection, transparency, and accountability elements and adapting to evolving legal frameworks are considered crucial for understanding the "decision making" process of AI.\cite{41}

\subsubsection{Transparency $\&$ Explainability}
It's essential to emphasize the significance of transparency and explainability in decisions made by AI, especially in areas like medical imaging where AI algorithms interpret intricate images.\cite{42} Explainability allows healthcare professionals 
to assess AI's decisions, enhancing confidence and fostering a better collaboration between AI systems, specialists,\cite{43} and patients who are directly affected by these decisions.\cite{44} Also, explainability empowers patients to make well-informed 
choices regarding their healthcare in AI-based diagnoses and treatments.\cite{45} Similarly, transparency enables healthcare practitioners to understand specific diagnoses or treatment prescriptions made by AI, as well as to audit AI systems to 
ensure they are functioning as intended, thus maintaining accountability.\cite{46} Legally, the matters of transparency and explainability in healthcare are intricate and continuously evolving. This complexity stems from legal concerns such as liability, 
accountability, and patient rights associated with the medical imaging sector. The primary challenge is to determine who to hold accountable if an AI system provides an incorrect diagnosis or treatment recommendation: the healthcare provider, 
the AI developer, or the AI itself. The lack of specific laws and regulations governing AI in healthcare further complicates these legal issues.\cite{31} Therefore, it is crucial to ensure transparency in the development and application of AI algorithms 
to identify and mitigate potential risks or errors.\cite{42}

\subsubsection{Accountability}
The concept of defining responsibility within the sphere of Artificial Intelligence (AI) presents significant challenges due to the inherent intricacies and lack of transparency of AI systems.\cite{36} To answer the question, "Where does the responsibility 
for AI-induced errors fall?" a thorough evaluation of legal liability, and notably, the aspect of responsibility, is necessitated. This is crucial as it underlines the demand for accountability and guarantees that appropriate measures are 
implemented in the event of damages.\cite{56} From a legal viewpoint, it could be proposed that developers who design the system in a manner that results in harm or fail to ensure its safety should bear the legal responsibility. Conversely, if users 
handle the AI system recklessly or improperly, they could be held accountable for negligence. However, if the AI system exhibits autonomous decision-making capacity and inflicts harm through its activities, it could be held responsible for 
those actions.\cite{57} The next logical question is how causation can be attributed to the non-human AI system. Firstly, it is essential to thoroughly comprehend the critical factors and processes that led to a particular outcome. Stakeholders should 
be able to determine whether the AI system's actions were intentional, accidental, or the result of concealed implicit biases or flaws, a concept closely tied to the principles of transparency and explainability. Secondly, the technical challenge 
lies in identifying a single user responsible for the damage incurred since algorithms, data sources, and human input form an integral part of AI systems. This complexity raises the question of the extent of responsibility that should be assigned 
to developers, users, or even the AI system itself.\cite{36}

\subsubsection{Data privacy $\&$ protection}
In the domain of medical imaging, artificial intelligence (AI) methodologies frequently require access to a broad range of patient information, encompassing both medical imagery and related health records. This becomes particularly essential when 
AI systems scrutinize and learn from considerable quantities of confidential patient data. Thus, the elements of data privacy and data protection within legal considerations demand that data collection, storage, and usage are carried out in compliance 
with the relevant data protection laws.\cite{61} The transfer and storage of sensitive patient information during the usage of AI in medical imaging amplify the threat of data breaches.\cite{63} And any unauthorized intrusion into such data can instigate harm and 
raise privacy issues.\cite{s27} Furthermore, the evaluation of medical imaging data by AI might risk patient re-identification, notwithstanding the confidentiality measures in place.\cite{65} Advanced techniques might enable de-identified data to be associated with 
other publicly accessible information, thereby endangering patient privacy.\cite{61} Also, AI algorithms employed in medical imaging may harbour biases,\cite{69} potentially leading to inconsistencies in patient care, erroneous diagnoses, or unequal treatment.\cite{44} 
This practice not only infringes on privacy by amplifying existing biases but also jeopardizes the confidentiality of certain groups.\cite{69}

\subsubsection{Intellectual property rights}
As previously highlighted, Intellectual Property Rights (IPR) can indeed intersect with, and occasionally conflict with, the need for transparency in legal aspects. Nevertheless, this area presents its unique set of challenges and plays a pivotal role 
in the advancement of novel algorithms and AI technologies in healthcare. These rights offer legal safeguards to the originators of such innovations, thus fostering innovation and investment in the field. By gaining exclusive rights to their creations, 
inventors may disclose and commercialise their inventions, which in turn propels progress in medical imaging and other healthcare applications.\cite{62} The matter becomes considerably more intricate when new algorithms and AI technologies are conceived by an 
AI without direct human intervention. Does this imply that the possession and entitlement to intellectual property are transferred to the AI? Conventionally, legal frameworks predominantly acknowledge human inventors as the legitimate owners of intellectual 
property. However, when it comes to AI-generated innovations, the issues of authorship and ownership are not straightforward. Some scholars have observed that resolving this dilemma necessitates meticulous scrutiny of existing IP laws, along with potential 
revisions to accommodate the distinct nature of AI-generated inventions.\cite{73} From a legal standpoint, if the allocation of ownership and entitlement to intellectual property is indeterminable, a significant challenge arises in establishing liability for 
AI-generated inventions and their potential infringement upon existing patents. Moreover, issues emerge concerning the extent of human participation in AI-generated inventions, the repercussions on conventional notions of creativity and inventiveness, 
and the demand for transparency and accountability in dealing with autonomous AI systems.\cite{74} The development of comprehensive legal frameworks that tackle these complexities is pivotal to stimulating innovation while safeguarding intellectual property rights.\cite{75} 
However, it's debatable how far the fundamental principles of intellectual property law and the intersection of privacy and intellectual property can accommodate these complexities. Some scholars propose that both human and AI contributions should be considered, 
along with the threshold of originality, as an approach to ascertain originality and copyright protection for AI-generated content.\cite{76} Others recommend a re-evaluation of the core principles and notions of intellectual property law,\cite{77} or suggest viewing 
privacy itself as a form of intellectual property.\cite{78} Overall, these discussions emphasize the intricacy of determining originality and copyright protection in AI-produced works, along with the implications of AI on legal aspects of intellectual property.

\subsubsection{Regulatory compliance}
The final aspect of legal considerations surrounding medical imaging pertains to the assessment of existing regulations in the UK and the element of regulatory compliance. Regulated and compliant AI systems is integral to ensuring safe and effective patient care, 
upholding ethical standards, and maintaining public trust. Therefore, aligning AI systems with existing regulations such as the Data Protection Act 2018 is of paramount importance. However, the swift evolution of AI technologies could potentially outpace these 
regulatory measures, necessitating continual updates to the legal framework.\cite{39} There are numerous other measures implemented to address transparency and explainability in AI, such as published guidelines and frameworks like the "Code of Conduct for Data-Driven 
Health and Care Technology" and the "NHS AI Lab AI Ethics Framework",\cite{80} are viewed as significant steps in addressing related concerns. These measures are perceived to serve as tools for transparency, explainability, and accountability, thereby maintaining public 
trust and ethical standards.\cite{price} Meanwhile, the Medicines and Healthcare Products Regulatory Agency (MHRA) supervises the regulation of AI and medical devices, including software. But there is an ongoing discourse regarding the efficiency of these regulations in 
keeping pace with the rapid advancement of AI and the complex issues it introduces.\cite{82} 

To tackle issues of accountability, the UK government has instituted the Centre for Data Ethics and Innovation (CDEI) to offer guidance on the ethical application of AI,\cite{39} and the protection of individuals' data rights.\cite{84} The Information Commissioner's Office 
(ICO)\cite{85} is another regulatory body tasked with enforcing data protection laws within the UK. They dispense guidance on data protection practices and probe into any instances of breaches or non-compliance.\cite{86} In contrast, the Intellectual Property Office (IPO) offers 
guidance concerning the patentability of AI inventions and the ownership of AI-produced works. Moreover, The Copyright, Designs and Patents Act 1988 has been revised to elucidate aspects related to AI-generated works.\cite{87} Nevertheless, the effectiveness and 
efficiency of these laws in dealing with the intricacies of AI and intellectual property necessitate continual appraisal and potential modification to keep abreast of technological progressions and legal challenges.\cite{88, 89, 90, 91}

In Appendix B, we provide more information on the legal considerations.

\subsection{Algorithmic and Technical aspects}
With respect to algorithmic aspects of the application of Generative AI in Healthcare, there are several ways to look at it from. This report looks at the subject from 6 angles namely: 

\begin{itemize}
    \item Unavailability of sufficient data for training. 
    \item Issues with accuracy and hallucinations. 
    \item Model training and compute limitations. 
    \item Issues with interpretability and explainability. 
    \item Issues with workflow integration. 
    \item The need for real-time inference and processing. 
\end{itemize}

\subsubsection{Unavailability of sufficient data for training}
The healthcare domain faces unique challenges in applying generative AI models due to limited availability of data, sensitive patient information, and scarcity 
of well-posed discriminative tasks. Additionally, the datasets used for benchmarking often fail to represent real-world data due to spectrum bias and measurement 
or imaging device discrepancies. To address these challenges, the AI for Health Imaging Initiative (AI4HI)\cite{dataunav2} was formed, bringing together five EU projects to develop 
Big Data infrastructure while adhering to GDPR. Individually, CHAIMELEON collects data in local or central databases, EuCAnImage builds a federated cancer imaging 
platform, INCISIVE aims for a pan-European repository of multi-modal data, ProCAncer-1 focuses on prostate cancer imaging, and PRIMAGE develops a cloud-based 
platform for pediatric cancer decision-making. Additionally, OpenAI has announced partnerships to produce public and private datasets for enhanced AI training. 

\subsubsection{Issues with accuracy and hallucination}
Hallucination refers to the generation by a LLM seemingly correct sentences by virtue of its innate probabilistic generation, which aren't entirely based on facts and can be largely inccorrect. This can lead to inaccurate and misleading outputs that could have serious consequences in healthcare. There are several sources of hallucinations in LLMs, some of which includes:
\begin{itemize}
    \item Unreliable sources of information: The training data for LLMs can be unreliable, such as if it contains false or inaccurate information. 
    \item Probabilistic generation: LLMs generate text based on probabilities, which means that there is a chance that they will generate incorrect or misleading content. 
    \item Biased training data: If the training data is biased, the model will be more likely to generate biased content. 
    \item Insufficient context: LLMs may not be able to understand the context of a query, which can lead to inaccurate or misleading responses. 
\end{itemize}
To measure and prevent hallucinations in, improve accuracy and foster workflow inclusion of LLMs, the following strategies are currently being used: 
\begin{itemize}
    \item Med-HALT\cite{medhalt}: A medical dataset framework designed to measure and evaluate hallucinations in LLMs. 
    \item Human evaluation\cite{medhalt2}: Having human experts evaluate the output of LLMs.
    \item Metrics\cite{medhalt2}: Using metrics such as perplexity and cross-entropy to assess the quality of LLM output.
    \item Human-in-the-loop: Having humans involved in the development and deployment of LLMs.
    \item Algorithmic corrections: Using machine learning techniques to correct hallucinations.
    \item Fine-tuning GPTs on healthcare data: Training GPT models on healthcare data. 
    \item Improving prompts: Using prompts that encourage the LLM to generate accurate and relevant content. Recently, OpenAI released this guide\cite{promptengineering} on prompt engineering.
    \item Instruction fine-tuning, as was reasonably effectively used in developing Google's QA LLM, FlanPaLM\cite{flanpalm} which achieved state-of-the-art performance in Medical Question and Answering tasks.
    \item The introduction of more robust medical datasets e.g. MultiMedQA, a new dataset which comprises 6 popular medical QA datasets (MedQA\cite{medqa}, MedMCQA\cite{medmcqa}, PubMedQA\cite{pubmedqa}, LiveQA\cite{LiveQA}, MedicationQA\cite{medicationqa}, MMLU Clinical Topics\cite{mmlu} and finally the newer HealthSearchQA\cite{healthsearchqa})
    \item Chain of prompt prompting\cite{chainofthought}: Using a series of prompts to help the LLM stay on track and avoid generating hallucinations. Commonly used alongside this in recent state-of-the-art Medical LLMs is self-consistency\cite{selfconsistency}.
    \item Retrieval-augmented generation (RAG)\cite{rag1}: Using vector databases to provide the LLM with additional context.
\end{itemize}

\subsubsection{Model training and compute}
There are challenges of training and deploying large language models (LLMs) in healthcare. One of the major challenges is the high cost of training LLMs, which can be prohibitive for 
healthcare facilities. The cost of training LLMs is due to the need for powerful hardware, such as GPUs, CPUs, and RAM storage. Additionally, data gathering and annotation 
is a time-consuming and labor-intensive process. To address these challenges, researchers have developed techniques such as model, data, and tensor parallelism to reduce 
the training time and optimize the compute requirements of LLMs. However, these techniques require effective communication between GPU nodes and high-speed internet 
connectivity, which may not be available to all healthcare facilities. Retrieval Augmented Generation (RAG), which is a technique that can be used to improve the performance of 
LLMs without the need for extensive training involving the use of vector databases to provide LLMs with additional context, can help them to generate more accurate and relevant 
responses could be a valuable tool for healthcare facilities that are looking to use LLMs without the high cost of training them. However, there is not yet much research into the application of RAG in healthcare. 

\subsubsection{Issues with intepretability and explainability}
In 2021, a study led by Michael Roberts et al.\cite{explainability1} highlighted issues in machine learning (ML) algorithms for COVID-19 detection from chest radiographs and CT scans. Algorithms wrongly associated disease severity 
with imaging views rather than features, hindering early-stage disease detection. In healthcare, minimizing wrong predictions and ensuring interpretable ML models is crucial. Vera Liao et al. research\cite{explainability2} on 
Large Language Models (LLMs) identified challenges in transparency due to complex architectures and proprietary technologies. Addressing interpretability, existing approaches include model reporting, evaluation 
result publication, explanation provision, and communicating uncertainty. Hallucinations in LLMs pose challenges, with research suggesting issues in training datasets and model reliance on memorization instead of 
robust learning. Open-source tools like EvidentlyAI\cite{evidentlyai} and Arize\cite{arize} aim to enhance LLM observability and interpretability by modifying prompts, tracking data changes, and fine-tuning models. However, the effectiveness of 
these strategies in improving trust and LLM adoption in healthcare requires further examination.

\subsubsection{Issues with workflow integration}
In addition to aforementioned issues such as hallucinations, inaccuracies and lack of transparency, healthcare practitioners (HCPs) draw insights from a diverse array of data sources, encompassing patient records, prescriptions, 
diagnostic images, and test results as such, single-mode LLMs i.e. those that deal with only text input, tend to have limited functionality in clinical settings for example. Recently, there has been the growing 
development of multi-modal LMMs comprising input from images/videos with text and/or sound input, the goal being to facilitate workflow integration and simulate the activities of physicians and other healthcare practitioners when 
making clinical decisions. Projections from a comprehensive EIT Health and McKinsey report\cite{eith} anticipate the automation of approximately 15$\%$ of healthcare workflows by 2030, potentially streamlining 20$\%$ to 8$\%$ of tasks performed 
by physicians and nurses through AI. Despite the considerable potential of AI in healthcare, its widespread adoption remains limited, with 44$\%$ of European healthcare professionals reporting minimal involvement in AI development and implementation. 
In this pursuit of broader AI integration in healthcare, there is an immediate need for innovative applications. For instance, the successful integration of Electronic Medical Records (EMR)\cite{brookings}, as highlighted in the 2022 Brookings report, 
underscores the pivotal role of innovation in software systems. Among the persisting challenges, the lackluster impact of existing AI models on workflow optimization is a notable concern for healthcare practitioners. Unimodal AI models, 
primarily employed in healthcare, particularly in radiology, fall short of significantly enhancing workflow. For instance, in cardiac diagnostics, cardiologists still heavily rely on traditional text-based medical records and lab reports in 
conjunction with medical images. In contrast, a comprehensive multimodal AI system adept at processing various data formats can liberate the cardiologist to focus on tasks demanding human expertise, representing a transformative shift in healthcare AI adoption.

\subsubsection{The need for real-time inference and processing}
Large Language Models (LLMs) and Large Multimodal Models (LMMs) pose several challenges when it comes to real-time inference in healthcare applications. These models, such as GPT-3\cite{gpt3} for LLMs and models like CLIP\cite{clip} for LMMs, have demonstrated 
impressive capabilities in understanding and generating human-like text and handling multimodal inputs. However, deploying them in real-time healthcare scenarios comes with specific challenges:

\begin{itemize}
    \item LLMs and LMMs are computationally expensive and resource intensive. Real-time inference requires quick processing, and these models may struggle to meet the time constraints of healthcare applications. The sheer size of the models and the \\
    number of parameters they possess make it challenging to deploy them on resource-constrained devices, like those commonly found in healthcare environments.
    \item Real-time healthcare applications demand low latency to provide timely responses. The inference time of LLMs and LMMs may be too high for certain use cases, impacting the overall responsiveness of the system. High latency can be critical 
    in healthcare scenarios where quick decision-making is essential, such as in emergency situations or during surgery.
    \item Storing and loading large models can be challenging in healthcare environments with limited storage capacity. Deploying such models on edge devices may require compromises in terms of model size and capabilities. Model updates and maintenance 
    can be cumbersome, especially if frequent updates are necessary to keep the model's knowledge up to date. 
    \item Healthcare data is highly sensitive, and using cloud-based services for real-time inference with LLMs and LMMs may raise concerns about data privacy and security. On-device deployment might mitigate some of these concerns, but it brings challenges \\
    related to the computational power and storage capacity of the devices. \\
\end{itemize} 
To address these challenges, ongoing research and development are necessary to optimize the efficiency, interpretability, and privacy aspects of LLMs and LMMs for real-time healthcare applications. Additionally, tailored training on healthcare-specific data and collaboration with domain experts are crucial to enhance the models' performance and relevance in medical contexts. 

For a more comprehensive detail on the algorithmic challenges, please visit Appendix C.

\section{Conclusion}
Generative AI in medical imaging and text analysis stands at a crossroads. While its potential to revolutionize healthcare is undeniable, navigating the ethical, legal, societal, and algorithmic complexities is paramount to ensure its responsible implementation. Addressing ethical concerns around accuracy, informed consent, and data privacy necessitates robust frameworks and ongoing public dialogue. The legal landscape needs clear 
regulations on liability and accountability, fostering innovation while safeguarding patient well-being. Societally, ensuring equitable access to AI-powered healthcare and mitigating potential biases within algorithms are crucial for building trust and inclusivity. Algorithmically, overcoming limitations in training data, model capabilities, and workflow integration requires continuous research and development, prioritizing 
real-time inference and explainability. By acknowledging these interconnected challenges and actively seeking solutions, we can unlock the transformative potential of generative AI in medicine, paving the way for a future where technology enhances human expertise, leading to a brighter horizon for healthcare and human well-being.

\appendices
\section{Societal aspects}
The advent of artificial intelligence (AI) technologies has revolutionized many sectors, including healthcare. Particularly, as mentioned, AI can be instrumental in medical imaging, enabling precision and accuracy in detecting and 
diagnosing diseases. However, this increased reliance on AI may impact the nature of social interaction in healthcare. 

Healthcare has traditionally been a high-touch field, with human interaction playing a critical role in patient care. The introduction of AI, while augmenting diagnostic efficiency, could potentially alter this dynamic. 
It may reduce direct interaction between healthcare providers and patients, as AI takes over tasks such as image interpretation and diagnosis. This could lead to a shift in the patient-provider relationship, posing questions 
about the importance of human touch and interaction in the healing process.\cite{s1} According to a study by Huang and Liang (2020), the potential lack of human touch in AI-based services may affect patient satisfaction and the 
perceived quality of care.\cite{s2} They argue that while AI improves efficiency, it may not be able to replicate the empathy, reassurance, and personal care that human healthcare providers offer.  Patients often derive comfort 
from the human interaction they receive during their healthcare experience, which AI, at its current state, cannot provide. This lack of 'human touch' could lead to a perceived decrease in the quality of care and may affect 
patient satisfaction. The implementation of AI in healthcare settings also brings forth complex societal issues that need addressing. These issues span the spectrum from ethical considerations, such as informed consent 
and privacy concerns, all of which has been already discussed in this paper, to social implications, like potential job displacement and the digital divide.\cite{s3} 

From a societal perspective, the increased use of AI could potentially displace jobs, particularly those that involve routine tasks. AI can automate certain tasks, freeing up radiologists and other imaging 
professionals to focus on areas requiring human expertise. For example, AI algorithms can be trained to accurately identify abnormalities in imaging scans, thus improving efficiency and reducing the workload of 
clinicians.\cite{s4} However, AI is also likely to create new roles within the field. The design, development, deployment and maintenance of AI systems will require new skills, leading to the creation of jobs such as AI 
specialists and data scientists.\cite{s5} This necessitates a shift in the education and training of healthcare workers. Reskilling will be key in helping these professionals adapt to the changing landscape. 
Medical education will need to incorporate AI literacy, including understanding the strengths and limitations of AI, data interpretation, and ethical considerations.\cite{s6} Therefore, continuous professional 
development programs will need to focus on skill acquisition in AI-related areas. Furthermore, interdisciplinary collaboration between medical professionals, computer scientists, and engineers will become 
increasingly important.\cite{s7} Efforts should also focus on developing AI applications that complement, rather than replace, the human aspects of healthcare. This could involve creating AI systems that still involve 
human healthcare providers in the decision-making process, or developing AI tools that assist healthcare providers in delivering more personalised care.\cite{30} 

Moreover, there's a risk of exacerbating the digital divide, as access to AI-powered healthcare may be limited to certain segments of society, potentially widening health disparities. 
It is accepted that AI algorithms can help analyse medical images more efficiently and accurately, potentially reducing diagnostic errors, enhancing patient care, and improving health outcomes.\cite{s9} 
However, if not properly managed, AI could also exacerbate health disparities. It's important to note that AI algorithms are often trained on datasets that might not be representative of diverse populations. 
If these datasets predominantly include data from certain groups and exclude others, this could lead to biased outcomes and widen existing health disparities.\cite{s10} Additionally, AI can potentially improve 
access to care in underserved areas by allowing for remote diagnostics and consultations, thus overcoming geographical barriers.\cite{s11} However, this opportunity also presents a risk: if AI advancements are primarily 
accessible to affluent communities due to cost or infrastructure requirements, it could increase health inequalities. Addressing health disparities in AI requires a careful consideration of legal frameworks. 
Issues such as privacy, data security, and informed consent are key.\cite{s12} Furthermore, regulatory mechanisms must ensure the quality and safety of AI systems and their algorithms. Anti-discrimination laws must also 
be enforced to prevent biased data collection and usage.\cite{s13} 

Lastly, while social aspect of AI is analysed, consideration of public trust is a crucial factor for the adoption of AI in healthcare, including in the realm of medical imaging. AI can potentially 
revolutionize healthcare by making diagnoses more accurate and efficient, but this relies on public acceptance and trust.\cite{s14} With the increasing use of AI in medical imaging, concerns about privacy, reliability, 
and fairness are paramount.\cite{s15} Public trust is the bedrock upon which the successful implementation and acceptance of AI applications in healthcare are built.16 Transparency about how AI is used in medical imaging is 
vital to gain and maintain public trust. This involves clear communication about the AI decision-making process, its accuracy, and potential limitations.\cite{s17} Regulations that protect patient interests are also important. 
This includes laws to safeguard patient data, ensure the reliability and validity of AI algorithms, and uphold the principles of fairness and equity.\cite{s18} Strong regulations thus serve as a guarantee to the public 
that their interests are protected, further fostering trust.\cite{s19} Societal frameworks play a fundamental role in fostering public trust in AI use in healthcare. These frameworks include educational initiatives to 
improve public understanding of AI, and mechanisms for public input and feedback on AI use.\cite{s20} Complex societal issues such as inequality and bias in AI algorithms must also be addressed. 
For instance, AI systems trained on data from one demographic might not work as well for another, leading to disparities in healthcare outcomes.\cite{s21} Policies should be in place to ensure diversity in training data and to test AI systems across different groups.\cite{s22} 
In the UK, AI in healthcare is regulated under several laws, including the Data Protection Act 2018, which governs the use of personal data, and the Human Rights Act 1998, which protects individuals.\cite{s23} While laws such as the 
GDPR provide a foundation for data protection, they are not explicitly designed to address the unique challenges posed by AI. For instance, GDPR's 'right to explanation' may not be fully applicable to complex AI models due to 
their 'black box' nature.\cite{s24} The National Health Service (NHS) has also developed an AI lab to ensure that AI technologies are safe and effective. These include the NHS Long Term Plan, which aims to make digital health services 
available to everyone, and the NHSX, an organization tasked with driving digital transformation in the NHS (NHS, 2019).\cite{s25} Resources such as funding and technical support are also provided to health tech innovators through initiatives 
like the AI in Health and Care Award (NHSX, 2021).\cite{s26} However, the effectiveness of these laws in ensuring equitable access to AI technologies in healthcare is still a subject of ongoing debate.\cite{s27} 

Lastly, the application of AI in medical imaging and consequent job displacement may raise a complex legal issue, particularly in respect to employment law. For instance, employment contracts may need to be revised to consider the potential for 
job role changes or redundancies due to AI implementation. The ethical implications of job displacement are also significant. The healthcare sector will need to ensure fair treatment of employees and consider the potential for discrimination, 
particularly if certain groups are disproportionately affected by job displacement (Dignum, 2017).\cite{s28} The UK government has recognized these challenges and launched a review into AI and the law led by the Law Commission in 2021.\cite{s29} 
Yet, it remains to be seen how effectively these future laws will tackle the issue of AI and patient experience in healthcare. 

\section{Legal aspects}
The incorporation of artificial intelligence (AI) into medical imaging brings immense potential to transform healthcare practices, promising enhanced diagnostic precision, improved patient outcomes, and streamlined healthcare delivery. 
Nevertheless, as the use of AI in medical imaging expands, it is crucial to address legal considerations linked with its application. To guarantee responsible and efficient use of AI, several key factors need thorough consideration. 
This section delves into the importance of factors such as accuracy, informed consent, transparency and explainability, accountability, data privacy, data protection, and intellectual property rights in the context of AI integration in medical imaging. 
Comprehending and integrating these vital elements is necessary to meet the legal stipulations that lay the foundation for successful AI integration into this crucial healthcare sector, while also protecting patient rights and fostering trust in the technology. 

\subsubsection{Accuracy}
Although AI is considered to have the potential to improve medical efficiency and precision, especially in medical imaging, its reliance on the quality of training data can lead to diagnostic errors, which could potentially harm patients\cite{30}. 
In conventional practices, incorrect diagnoses can result in legal liabilities, with the responsibility usually resting on the healthcare providers. However, in the context of AI, an important question emerges: "Who is accountable for AI negligence - the healthcare providers, the developers, or the institutions using AI?"\cite{31}
Considering AI's self-governing decision-making and learning capabilities could reshape conventional legal notions of liability and accountability, it is debated that distinctive legal frameworks need to be established\cite{32}. From a legal viewpoint, accuracy isn't solely about a correct diagnosis; 
it also relates to the rights and dignity of those availing these services\cite{33}.

\subsubsection{Informed consent}
One potential solution to eliminate this issue is to provide patients with an informed choice, making them aware of potential risks and benefits of proposed medical procedures\cite{34}. Based on this informed consent, patients have the autonomy to make their own decisions, 
a principle that is legally mandated in healthcare systems globally\cite{35}. The significance of informed consent escalates when patient data is processed by AI\cite{36}, since decisions or recommendations made by AI systems based on this data can significantly impact patient health, 
underscoring the critical nature of informed consent.\cite{37} However, the concept of informed consent poses a considerable challenge due to the complexities of understanding AI systems. Some patients may struggle to grasp the potential risks and benefits of AI integration in their medical procedures and 
the ‘decision making’ process of AI. This lack of understanding could potentially undermine the legitimacy of informed consent\cite{38}. Understanding and defining the "decision making" process also brings forth complex legal concerns. This requires addressing issues such as data protection, transparency, and accountability\cite{39}. 
However, the 'black box' nature of AI, where decision-making processes are not easily understood, adds another layer of complexity\cite{40}. 
Therefore, incorporating data protection, transparency, and accountability elements and adapting to evolving legal frameworks are considered crucial for understanding the "decision making" process of AI\cite{41}.

\subsubsection{Transparency and Explainability}
From the above observations, it’s essential to emphasize the significance of transparency and explainability in decisions made by AI, especially in areas like medical imaging where AI algorithms interpret intricate images\cite{42}. 
Explainability allows healthcare professionals to assess AI's decisions, enhancing confidence and fostering a better collaboration between AI systems, specialists\cite{43}, and patients who are directly affected by these decisions\cite{44}. 
Also, explainability empowers patients to make well-informed choices regarding their healthcare in AI-based diagnoses and treatments\cite{45}. Similarly, transparency enables healthcare practitioners to understand specific diagnoses or treatment prescriptions made by AI, as well as to audit AI systems to ensure they are functioning as intended, 
thus maintaining accountability\cite{46}. Legally, the matters of transparency and explainability in healthcare are intricate and continuously evolving. This complexity stems from legal concerns such as liability, accountability, and patient rights associated with the medical imaging sector. 
The primary challenge is to determine who to hold accountable if an AI system provides an incorrect diagnosis or treatment recommendation: the healthcare provider, the AI developer, or the AI itself. The lack of specific laws and regulations governing AI in healthcare further complicates these legal issues\cite{47}. Therefore, 
it is crucial to ensure transparency in the development and application of AI algorithms to identify and mitigate potential risks or errors\cite{48}.

Researchers like Lundberg and Lee (2017)\cite{49}, Holzinger et al. (2017)\cite{50}, Caruana et al. (2015)\cite{51}, and Montavon et al. (2018)\cite{52} have contributed significant findings to the discussion on explainability and transparency in AI. They stress the importance of creating interpretable AI models, model-agnostic interpretability techniques, visual aids, 
uncertainty assessment, and algorithm documentation to enhance the transparency and explainability of medical imaging by AI. However, when intellectual property is concerned, transparency and explainability can pose challenges. AI algorithms may be proprietary, making it hard to offer complete transparency without revealing the entire algorithm. 
AI developers often strive to safeguard their algorithms and models through intellectual property rights such as patents, copyrights, or trade secrets. Therefore, striking a balance between intellectual property protection and transparency is essential\cite{53}. Legal frameworks must consider public interest, patient safety, and ethical considerations 
in healthcare. Regulations can be developed that require AI developers to provide explanations or justifications for AI decisions without disclosing the entire algorithm, promoting transparency and accountability while respecting intellectual property rights\cite{54}.

\subsubsection{Accountability}
The concept of defining responsibility within the sphere of Artificial Intelligence (AI) presents significant challenges due to the inherent intricacies and lack of transparency of AI systems\cite{55}. To answer the question, "Where does the responsibility for AI-induced errors fall? Is it on the shoulders of healthcare providers, developers, 
or institutions employing AI?" a thorough evaluation of legal liability, and notably, the aspect of responsibility, is necessitated. This is crucial as it underlines the demand for accountability and guarantees that appropriate measures are implemented to rectify any inflicted harm. Responsibility in the AI domain encompasses both ethical and legal dimensions. 
Ethically, it is incumbent upon developers and users to ensure that AI systems are devised and used ethically, considering all potential risks and implications. From a legal standpoint, accountability facilitates in ascertaining liability and apportioning repercussions in the event of damages. Nonetheless, determining liability for damages resulting from an 
AI system is a complex task that relies on an array of factors. The developer, the user, and the AI system itself could all bear various levels of accountability depending on their roles and actions. From a legal standpoint, accountability facilitates in ascertaining liability and apportioning repercussions in the event of damages.\cite{56} 
Nonetheless, determining liability for damages resulting from an AI system is a complex task that relies on an array of factors. The developer, the user, and the AI system itself could all bear various levels of accountability depending on their roles and actions. From a legal viewpoint, it could be proposed that developers who design the system in a manner that results in harm or fail to ensure its safety should bear the 
legal responsibility. Conversely, if users handle the AI system recklessly or improperly, they could be held accountable for negligence. However, if the AI system exhibits autonomous decision-making capacity and inflicts harm through its activities, it could be held responsible for those actions\cite{57}. 

The complexity of this task may not be readily apparent to the layperson. The autonomous nature of Artificial Intelligence (AI) systems, their capacity for independent decision-making, and the intricacies involved in determining liability and causation, all contribute to significant legal challenges. 
Although laws regulating AI and liability are continually adapting to accommodate its unique attributes, it remains crucial to address these distinct characteristics when attributing causation to an AI system. This attribution must incorporate concepts such as strict liability, negligence, and product liability\cite{58}. Therefore, the next 
logical question concerns how causation can be attributed to the non-human AI system. Firstly, it is essential to thoroughly comprehend the critical factors and processes that led to a particular outcome. Stakeholders should be able to determine whether the AI system's actions were intentional, accidental, or the result of concealed implicit biases or flaws, 
a concept closely tied to the principles of transparency and explainability. Secondly, the technical challenge lies in identifying a single user responsible for the damage incurred since algorithms, data sources, and human input form an integral part of AI systems. This complexity raises the question of the extent of responsibility 
that should be assigned to developers, users, or even the AI system itself\cite{59}.

\subsubsection{Data privacy and Data protection}
In the domain of medical imaging, artificial intelligence (AI) methodologies frequently require access to a broad range of patient information, encompassing both medical imagery and related health records. This becomes particularly essential when AI systems scrutinize and learn from considerable quantities of confidential patient data. 
Thus, the elements of data privacy and data protection within legal considerations demand that data collection, storage, and usage are carried out in compliance with the relevant data protection laws\cite{60}. Thus, the elements of data privacy and data protection within legal considerations demand that data collection, storage, and usage are 
carried out in compliance with the relevant data protection laws\cite{61}. In addition, it is advisable, when possible, to present patients with precise and extensive details about the purpose of collecting their data and the potential risks linked to the application of their information by AI in medical imaging. As a result, their agreement to the use of 
their personal data should be based on this knowledge, in conjunction with the protections established to preserve their privacy\cite{62}. 

While the utilization of artificial intelligence (AI) in medical imaging holds promise for elevating diagnostic precision and enhancing patient results, it simultaneously introduces novel privacy-related legal hurdles. Non-compliance with data privacy regulations often leaves patients susceptible to various violations. 
The transfer and storage of sensitive patient information during the usage of AI in medical imaging amplify the threat of data breaches\cite{63}. And any unauthorized intrusion into such data can instigate harm and raise privacy issues\cite{64}. Furthermore, the evaluation of medical imaging data by AI might risk patient re-identification, notwithstanding the 
confidentiality measures in place\cite{65}. Advanced techniques might enable de-identified data to be associated with other publicly accessible information, thereby endangering patient privacy\cite{66}. Also, AI algorithms employed in medical imaging may harbour biases\cite{67}, potentially leading to inconsistencies in patient care, erroneous diagnoses, or unequal treatment\cite{68}. 
This practice not only infringes on privacy by amplifying existing biases but also jeopardizes the confidentiality of certain groups\cite{69}.

Moreover, the fact that AI systems might be designed and managed by third parties, potentially resulting in inappropriate handling or misuse of patient data, compounds the problem\cite{70}. It is worth noting that informed consent from patients is vital for the application of AI in medical imaging. However, there could be instances 
where consent procedures are insufficient, leading to potential privacy violations and legal implications\cite{71}.

\subsubsection{Intellectual property rights}
As previously highlighted, Intellectual Property Rights (IPR) can indeed intersect with, and occasionally conflict with, the need for transparency in legal aspects. Nevertheless, this area presents its unique set of challenges and plays a pivotal role in the advancement of novel algorithms and AI technologies in healthcare. 
These rights offer legal safeguards to the originators of such innovations, thus fostering innovation and investment in the field. By gaining exclusive rights to their creations, inventors may disclose and commercialise their inventions, which in turn propels progress in medical imaging and other healthcare applications\cite{72}. The matter becomes 
considerably more intricate when new algorithms and AI technologies are conceived by an AI without direct human intervention. Does this imply that the possession and entitlement to intellectual property are transferred to the AI? Conventionally, legal frameworks predominantly acknowledge human inventors as the legitimate owners of intellectual 
property. However, when it comes to AI-generated innovations, the issues of authorship and ownership are not straightforward. Some scholars have observed that resolving this dilemma necessitates meticulous scrutiny of existing IP laws, along with potential revisions to accommodate the distinct nature of AI-generated inventions\cite{73}. From a legal standpoint, 
if the allocation of ownership and entitlement to intellectual property is indeterminable, a significant challenge arises in establishing liability for AI-generated inventions and their potential infringement upon existing patents. Moreover, issues emerge concerning the extent of human participation in AI-generated inventions, the repercussions on 
conventional notions of creativity and inventiveness, and the demand for transparency and accountability in dealing with autonomous AI systems\cite{74}. The development of comprehensive legal frameworks that tackle these complexities is pivotal to stimulating innovation while safeguarding intellectual property rights\cite{75}. However, it's debatable how far the 
fundamental principles of intellectual property law and the intersection of privacy and intellectual property can accommodate these complexities. Some scholars propose that both human and AI contributions should be considered, along with the threshold of originality, as an approach to ascertain originality and copyright protection for AI-generated content\cite{76}. 
Others recommend a re-evaluation of the core principles and notions of intellectual property law\cite{77}, or suggest viewing privacy itself as a form of intellectual property\cite{78}. Overall, these discussions emphasize the intricacy of determining originality and copyright protection in AI-produced works, along with the implications of AI on legal aspects of intellectual property.

\subsubsection{Regulatory compliance}
The final aspect of legal considerations surrounding medical imaging pertains to the assessment of existing regulations in the UK and the element of regulatory compliance. Regulated and compliant AI systems is integral to ensuring safe and effective patient care, upholding ethical standards, and maintaining public trust. 
Therefore, aligning AI systems with existing regulations such as the Data Protection Act 2018, which conforms to the European General Data Protection Regulation (GDPR), is of paramount importance. However, the swift evolution of AI technologies could potentially outpace these regulatory measures, necessitating continual updates to the legal framework\cite{79}.

There are numerous other regulations, guidelines, and acts designed to address the various legal aspects of AI application, with some focusing specifically on the application of AI systems in healthcare. Measures implemented to address transparency and explainability in AI, such as published guidelines and frameworks 
like the "Code of Conduct for Data-Driven Health and Care Technology" and the "NHS AI Lab AI Ethics Framework"\cite{80}, are viewed as significant steps in addressing related concerns. These measures are perceived to serve as tools for transparency, explainability, and accountability, thereby maintaining public trust and ethical standards\cite{81}. 
Meanwhile, the Medicines and Healthcare Products Regulatory Agency (MHRA) supervises the regulation of AI and medical devices, including software. The MHRA is encouraged to take an initiative-taking stance in monitoring and regulating AI technologies. Notwithstanding, there is an ongoing discourse regarding the efficiency of these regulations in 
keeping pace with the rapid advancement of AI and the complex issues it introduces\cite{82}.

To tackle issues of accountability, the UK government has instituted the Centre for Data Ethics and Innovation (CDEI) to offer guidance on the ethical application of AI\cite{83}, and the protection of individuals' data rights\cite{84}. The Information Commissioner's Office (ICO)\cite{85} is another regulatory body tasked with enforcing data protection laws within the UK. 
They dispense guidance on data protection practices and probe into any instances of breaches or non-compliance\cite{86}. In contrast, the Intellectual Property Office (IPO) offers guidance concerning the patentability of AI inventions and the ownership of AI-produced works. 
Moreover, The Copyright, Designs and Patents Act 1988 has been revised to elucidate aspects related to AI-generated works\cite{87}. Nevertheless, the effectiveness and efficiency of these laws in dealing with the intricacies of AI and intellectual property necessitate continual appraisal and potential modification to keep abreast of 
technological progressions and legal challenges.\cite{88,89,90,91}  Despite endeavours to regulate, protect, and enforce these regulations, an existing legal framework faces significant challenges due to the unique characteristics of AI, such as its capacity for constant learning and adaptation. 
Additionally, ethical considerations like bias and fairness must also be legally addressed to ensure the just and equitable utilisation of AI\cite{92}. 

\section{Algorithmic/Technical aspects}
\subsubsection{Unavailability of sufficient data for training}
Generative AI models as with other machine learning models require the availability of large quantities of datasets regardless of the mode be it visual, audio or text. With generative models, their output depends on proper learning of the probabilistic and statistical relationships in the data they are trained with. General GPT models are usually trained on data of millions and billions of magnitude unlike other domains, 
the domain of healthcare is severely challenged with lack of data and even where available, there are issues with respect to the inclusion of sensitive patient information, classified under the two subclasses: Protected Health Information (PHI) and Personally Identifiable Information (PII). 
Compared to other domains, available data in healthcare is usually smaller and in the order of hundreds to thousands unlike other domains where data runs into millions. Also, there is the challenge in healthcare where very few clinical questions come as well-posed discriminative tasks, which slow their application in Machine Learning tasks. 
There is also the challenge of the dataset being used to test models for benchmarking being unrepresentative of real-world data; this can be caused by a number of factors including a test data selection cohort not appropriately representing the right range of possible symptoms, sometimes called spectrum bias, also issues with difference in imaging devices or measurement bias in procedures. 
This was very well emphasized in the case whereby almost hundred models were developed for the detection of COVID-19 but none of them proved useful in clinical settings. Other avenues of failure include the use of metrics during model development that do not matter in the real-world, for example model accuracy.\cite{dataunav1} 

In the European Union, to tackle the challenge of health data availability, a number of techniques have been put in place, one of which is the development of the AI for Health Imaging Initiative (AI4HI) which involves the coming together of 5 EU projects (CHAIMELEON, EuCAnImage, INCISIVE, ProCAncer-1, PRIMAGE) 
towards the development of Big Data infrastructure, while also taking cognizance of GDPR. Relevant points in this strategy involve the adoption of the Digital Imaging $\&$ Communications On Medicine (DICOM) as the data modeling standard, the use of centralized and/or federated data infrastructures. 
Individually; CHAIMELEON collects data which is stored in either local or central databases. These DICOM images are pseudonymized, where direct identifiers are either removed or replaced with randomly generated pseudonyms, 
EuCAnImage is building a federated European cancer imaging platform, comprising both local data warehouses and a central repository with anonymized data, 
INCISIVE aims to develop a pan-European federated repository of multi-modal data sources including imaging, biological and other non-imaging clinical data, unlike the others, INCISIVE builds on the FHIR (Fast Healthcare Interoperability Resources) and 
DICOM standard, the former which is a standard for healthcare data exchange, the ProCAncer-1 strategy aims to create a scalable, sustainable, quality-controlled prostate-cancer related medical imaging platform focused on answering a number of prostate-cancer related clinical questions, finally PRIMAGE aims to to develop a cloud-based platform to support decision-making in two main pediatric cancers: neuroblastoma and diffuse 
intrinsic pontine glioma. Similar to ProCAncer-1, PRIMAGE adopts a centralized architecture storing completely anonymized health data.\cite{dataunav2} 

Lastly, tech giant OpenAI recently announced its intention to partner \cite{dataunav3} with organizations towards producing public and private datasets for improved AI training. 

\subsubsection{Issues with accuracy and hallucinations}
Generative models particularly of the textual type have enjoyed significant devotion in terms of research and development with numerous LLMs currently in circulation however there exists any use of these models in domains where factual accuracy, legal risk and context-aware generation matters for example healthcare domains. 
Healthcare practitioners are currently very wary of applying LLMs in healthcare due to concerns of inaccurate generated outputs and hallucinations. Hallucinations as defined by popular LLM, ChatGPT-3.5 is “the generation of content that is not based on real or existing data but is instead produced by a machine learning model's extrapolation or creative interpretation of its training data. 
These hallucinations can manifest in various forms, such as images, text, sounds, or even video. AI hallucinations occur when a machine learning model, particularly deep learning models like generative models, tries to generate content that goes beyond what it has learned from its training data. These models learn patterns and correlations from the data 
they are trained on and attempt to produce new content based on those patterns. However, in some cases, they can generate content that seems plausible but is actually a blend of various learned elements, resulting in content that might not make sense or could even be surreal, dream-like, or fantastical”\cite{hallucinations1}. Sources of hallucinations include: 
\begin{itemize}
    \item Unreliable sources of information: where the data used in the training of the LLM proves to be unreliable. This is a common factor with all Generative Pre-Trained (GPT) models as they are trained on data which is available in the public domain which can many times be unverified and unreliable. 
    \item Probabilistic generation: as mentioned earlier, LLMs as with all generative models learn the statistical probabilities of information in their training data and use this information to generate outputs, also based on probabilities, this can result in cases where the model generates output which is probabilistically relevant to the input query but missing in context. 
    \item Biased training data: this results from situations where the data used to train the model was intentionally or unintentionally biased, some ways of generating bias is the use of wrong information or poor management of data, all of which lead to generation of incorrect outputs by LLMs.
    \item Insufficient context: LLMs are usually stateless, meaning unless actively done, they are sometimes unable to decipher the relevant context to the input query, which often results in the generation of outputs which might be relevant in one context but irrelevant to the context in which the query is asked. Tangentially related to this is scenarios where the LLM fails to take cognizance of salient information providing relevant context in the input data, latest research \cite{hallucinations2} has shown the fallibility in LLMs whereby they are unable to keep track and take note of information placed deep within a user query.
    \item Intrinsic nature: hallucination by LLMs can in a way be seen as a feature as opposed to a bug. When using LLMs, the typical temperature setting is greater than 0 which causes the model to become more creative, while this is beneficial as we do not want repeated responses, it also pre-disposes the model to generally factually incorrect responses.
\end{itemize}

Measuring hallucinations in LLMs is a necessary step in their development in order to ensure adoption in healthcare settings. To this end, a number of techniques have been developed some of which include the following: 
\begin{itemize}
    \item Med-HALT\cite{medhalt}: Medical Domain Hallucination Test for LLMs (Med-HALT) is a dataset framework designed to measure and evaluate hallucinations in LLMs. The framework comprises of datasets divided into 2 categories of tests: the first measuring the ability of the LLM to reason about a given problem; achieved through the use of False Confidence tests (FCT), None Of The Above tests (NOTA), Fake Questions Test (FQT), the second measuring the LLM’s ability to retrieve accurate information from its encoded training data. The tests feature questions from various countries and disciplines introducing variability in its testing process. 
    \item Human evaluation\cite{medhalt2}: where the response of LLMs are broken into atomic subsets which are then evaluated by experienced human evaluators, in the case of healthcare, this would mean physicians, pharmacists, pathologists etc.  
    \item Metrics such as perplexity and cross-entropy exist for LLMs, ROUGE also exists for sentence summarization, BLUE or Meteor for Machine Translation tasks\cite{medhalt2}. 
\end{itemize}

However, it should be mentioned that these techniques aren't fool-proof but they definitely contribute to the measurement of hallucinations in LLMs. Therefore, there have been the development of techniques for the mitigation of hallucinations ab-initio, some of which include: 
\begin{itemize}
    \item Human-in-the-loop: which involves having humans with domain expertise being actively involved in all aspects of machine learning development life cycle. Tangentially to this is the technique of Reinforcement Learning through Human Feedback (RLHF) \cite{rlhf} which is a technique using the reinforcement learning technique where the LLM is trained using human feedback collection and reward modelling to penalize itself for wrong outputs towards ensuring it eventually trains itself to generate accurate outputs.
    \item Analogous to the aforementioned and to eliminate the need for human feedback is techniques such as Reinforcement Learning from AI Feedback (RLAIF)\cite{rlaif} and Direct Preference Optimization (DPO); which though similar to RLHF is easier to implement as it optimizes a classification loss directly on preference data. A recent paper by uses DPO with promise.
    \item Algorithmic corrections using traditional machine learning techniques such as regularization, penalty loss, self-consistency\cite{selfconsistency} which uses an idea similar to ensemble techniques in machine learning. 
    \item Fine-tuning GPTs on healthcare data: while this seems like an attractive idea, it is a computationally expensive endeavor and not a perfect guarantee. 
    \item Improving prompts: this involves a number of techniques for example using phrases like “Answer as truthfully as possible, and if you're unsure of the answer, say "Sorry, I don't know” which forces the model to only output it is highly sure of. Similar to this, techniques such as the injection of knowledge graphs\cite{knowledgegraph} during prompting describe the use of data repositories interconnecting vast amounts of information in the form of entities and relationships. This helps mitigate hallucinations by mapping contextually relevant knowledge entities to the LLM's prompts. Because LLMs are trained in an unsupervised manner, there is minimal control over their outputs, with knowledge graph injection, one is able to implement controllable text generation with LLMs ensuring the generated text aligns with predefined attributes. 
    \item Related to the aforementioned is the idea of chain of prompt prompting: LLMs are known to struggle with multi-step reasoning whereby when tasked with providing answers to queries that involve methodical processing of different steps, they tend to struggle and can potentially hallucinate. To tackle this, chain-of-thought prompting\cite{chainofthought} is an idea whereby the user of the LLM asks a series of sequential questions from the LLM, making recourse to the answers from the previous queries, thus helping the model stay in state and prevent hallucinations. Similar research on this is Tree of Thought\cite{treeofthought}, Thread of Thought\cite{threadofthought}.  
    \item Another technique is the use of vector databases storing recent and relevant information on a subject matter or domain. Retrieval Augmented Generation (RAG)\cite{rag1, rag2} refers to the use of these vector databases which contain indexed embedding vectors of information on a subject matter and when a user supplies a query, a similarity mapping is done between the user query and the embedded information in the database, the best mapping is then combined with the query as input to the LLM. While it has its advantages, it is not an absolute solution as there exists issues of obtaining wrong contexts, lack of groundness\cite{grounded}; a situation whereby the response of the LLM is more influenced by its training data than the information obtained through retrieval, finally, even if the LLM retrieves right context, it may not completely answer the question query. 
\end{itemize}

\subsubsection{Model training and compute limitations}
Training state-of-the-art models in machine learning, particularly generative models demand high compute per year. Research\cite{compute} by generative AI giant OpenAI, shows a 10-fold increase in compute per year. It is a given that the more compute expended in training model, the better its performance. 
Hardware requirements for training LLMs in particular, can be demanding for organizations seeking to deploy LLMs for use, some of the hardware components which include large GPUs, CPUs, RAM storage are beyond the reach of the average healthcare facility. Training GPT model LLaMa\cite{llama} took 21 days to train requiring 2048 GPUs, 
costing approximately \$1M, OpenAI’s GPT-4\cite{gpt4} model costed more than \$100M to train, Meta's OPT\cite{opt} model required 992 80 Gigabyte Nvidia A100 GPUs, these are expenses beyond the reach of non-Big Tech companies. In addition to the hardware and financial requirements needed to train generative models, data gathering and annotation is a 
humongous task for healthcare companies to undertake, to tackle these companies are saddled with either employing staff tasked for this process or hiring external entities to make this happen, an approach employed by OpenAI\cite{dataannotation}. To reduce the training time and optimize the compute requirements in training LLMs, techniques such as model, 
data and tensor parallelism have been employed to make this happen. Model parallelism refers to the duplication of a given model across different GPUs, data parallelism involves the splitting of the training data into micro batches which are then distributed across different workers, tensor parallelism refers to the splitting of matrix multiplications across 
different GPUs. These techniques, though efficient, require effective communication between GPU nodes with fast internet connectivity and distributed networking setups, which might not be within the reach of healthcare facilities.  

The introduction of research augmented generation however allows for possible mitigation in training costs as healthcare facilities can opt to use LLMs. There hasn't been much research into the application of RAG-enhanced LLMs in healthcare, the Almanac\cite{almanac} research offers some possibilities.

\subsubsection{Issues with interpretability and explainability}
In 2021, \cite{explainability1} conducted a study on the pitfalls of machine learning algorithms in the detection of COVID-19 using chest radiographs and CT scans, they found instances where algorithms associated severe diseases not with image features but more with the view with which the image was taken. 
For example, for patients who are sick and immobile, the anteroposterior rather than posteroanterior view was used when taking images, consequently, deep learning models trained on these images learned wrong correlative relationships between the position of the image and the severity of the disease. This is obviously disadvantageous as 
it means the models are unable to detect diseases which are early stage and are taken posteroanterior.

In healthcare, it is pertinent that wrong predictions are brought to as minimal as possible, and it is equally important that ML models being used are easily interpretable. Model interpretability refers to the degree to which a human can understand the cause of a decision. \cite{explainability2} in their research 
into transparency with LLM use identified key challenges with achieving transparency with LLMs; the complex and uncertain behavior and capability of LLMs, the massive and opaque architectures used in their development (LLMs typically are developed using attention transformer framework which being a deep learning model makes it intrinsically hard to understand), 
the proprietary nature of LLM technologies which makes it hard for outsiders to study (although, efforts are being made at developing open-source LLMs), the use of LLMs through LLM-infused applications such as specialized chatbots and other productivity tools which will cause the opacity of the LLMs to trickle down to the applications built on them. Existing approaches to address explainability 
in LLMs include model reporting, publishing evaluation results, providing explanations (people achieve this by asking the LLMs to provide backings behind its results), the communication of uncertainty whereby the LLM gets to return its confidence level in the answer it provides.  

As mentioned earlier, hallucinations remain a big challenge with the adoption of LLMs in healthcare, recent research\cite{explainability3} indicates that hallucination stems from problems with the dataset on which the LLM was trained on and in the model itself whereby they still rely on memorization at sentence level and statistical patterns at corpora level instead of robust learning. 
With respect to dataset-related causes, the problem can be subclassified into the lack of relevant data and the duplication of data. Consequently, mitigation can be achieved through the improvement of the datasets and optimization of the LLM models e.g. through scaling. 

Currently, open-source tools are being developed to assist with observability $\&$ interpretability in LLMs. EvidentlyAI\cite{evidentlyai} for example helps modify prompts for accurate results, tracking changes to input data embeddings for specialized LLMs towards catching cases like data and concept drift. Another company, Arize\cite{arize} provides LLM observability and interpretability 
tools in the form of detecting problematic prompts, improving prompts through prompt engineering and fine-tuning LLMs using vector similarity search. However, it remains to be seen how effective their strategies are and what impact they will have on improving trust and adoption of LLMs in healthcare. 

\end{document}